\documentclass[sigconf]{acmart}

\AtBeginDocument{%
  }

\setcopyright{acmlicensed}
\copyrightyear{2024}
\acmYear{2024}
\acmDOI{}

\acmConference[AMC-ICAIF '24]{Make sure to enter the correct
  conference title from your rights confirmation emai}{Nov 14--17,
  2024}{Brooklyn, NY}
\acmISBN{}

\usepackage{bm}
\usepackage{multirow}
\usepackage{algorithm}       
\usepackage{algpseudocode}   
\usepackage{makecell} 
\usepackage{threeparttable}

\begin{document}
\title{Multi-Reranker: Maximizing performance of retrieval-augmented generation in the FinanceRAG challenge}

\author{Joohyun Lee}
\affiliation{%
  \institution{Financial Security Institute}
  \country{Republic of Korea}}
\email{dlee110600@gmail.com}

\author{Minji Roh}
\affiliation{%
  \institution{Financial Security Institute}
  \country{Republic of Korea}}
\email{nohmin0710@gmail.com}

\renewcommand{\shortauthors}{Lee et al.}

\begin{abstract}
As Large Language Models (LLMs) increasingly address domain-specific problems, their application in the financial sector has expanded rapidly. Tasks that are both highly valuable and time-consuming, such as analyzing financial statements, disclosures, and related documents, are now being effectively tackled using LLMs. This paper details the development of a high-performance, finance-specific Retrieval-Augmented Generation (RAG) system for the ACM-ICAIF '24 FinanceRAG competition. We optimized performance through ablation studies on query expansion and corpus refinement during the pre-retrieval phase. To enhance retrieval accuracy, we employed multiple reranker models. Notably, we introduced an efficient method for managing long context sizes during the generation phase, significantly improving response quality without sacrificing performance. We ultimately achieve 2nd place in the FinanceRAG Challenge. Our key contributions include: (1) pre-retrieval ablation analysis, (2) an enhanced retrieval algorithm, and (3) a novel approach for long-context management. This work demonstrates the potential of LLMs in effectively processing and analyzing complex financial data to generate accurate and valuable insights. The source code and further details are available at \url{https://github.com/cv-lee/FinanceRAG}.
\end{abstract}

\ccsdesc[500]{Computing methodologies~Natural language generation}
\keywords{Retrieval Augmented Generation, Large Language Model, Finance Analysis, Retrieval, Rerank}
\begin{teaserfigure}
  \includegraphics[width=\textwidth]{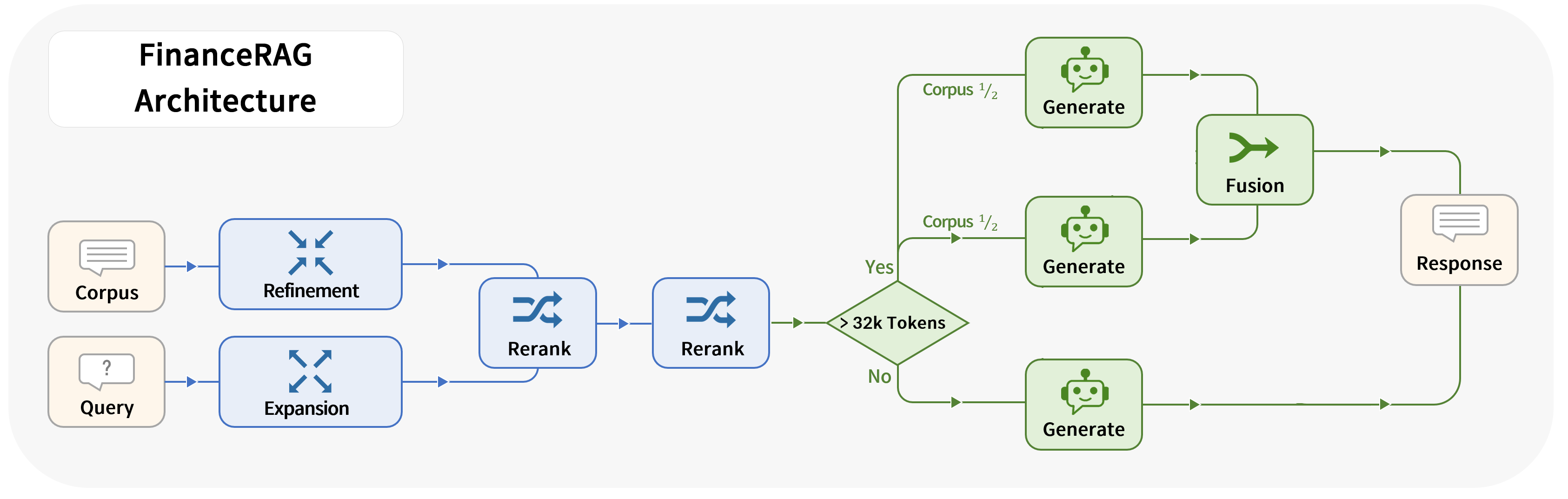}
  \caption{Overview of Retrieval Augmented Generation System}
  \Description{}
  \label{fig:teaser}
\end{teaserfigure}

\maketitle

\section{Introduction}

The advent of Large Language Models (LLMs) has revolutionized the approach to solving domain-specific problems, leading to a surge in their application within the financial sector \cite{setty2024improving, zhang2023enhancing, yepes2024financial}. Tasks that are both high in market value and traditionally time-consuming, such as analyzing financial statements, disclosures, press releases, and related documents, are now being actively addressed using LLMs \cite{setty2024improving}. Historically, the process of sifting through hundreds of pages to extract necessary information for investment decision-making was not only labor-intensive but also time-consuming, often causing delays that could result in financial losses. LLMs are effectively mitigating these issues by efficiently extracting relevant paragraphs from extensive documents and isolating the required information. \cite{gao2024modular, yepes2024financial}

In this paper, we present the development of a high-performance, finance-specific Retrieval-Augmented Generation (RAG) system, detailing our solution approach for the ACM-ICAIF '24 FinanceRAG competition task \cite{icaif-24-finance-rag-challenge}. We optimized our system through ablation studies of various techniques, including query expansion and corpus optimization during the pre-retrieval phase. To push the retrieval performance to its limits, we employed a combination of multiple reranker models. Moreover, we designed an innovative method for efficiently managing long context sizes during the generation phase, which significantly elevated the quality and accuracy of the responses. This approach allowed us to harness the full potential of LLMs even when dealing with exceedingly lengthy input contexts, thereby maximizing answer performance. Our contributions in this paper are:

\begin{itemize}
\item {\textbf{Pre-Retrieval Ablation Study}} : We conducted comprehensive ablation studies to optimize pre-retrieval techniques, enhancing the overall retrieval effectiveness.
\item {\textbf{Accurate Retrieval Algorithm}} : We developed an accurate retrieval algorithm by leveraging multiple reranker models, improving the relevance of the retrieved corpora.
\item {\textbf{Efficient Context Size Management}} : We designed a method to efficiently manage context sizes, enabling the processing of very long inputs without sacrificing performance.
\end{itemize}

\section{Task and Dataset}

To build a high-performance RAG system in the financial domain, we addressed two primary tasks as defined by the FinanceRAG competition at ACM-ICAIF '24 \cite{icaif-24-finance-rag-challenge}. Task 1 involves retrieving the top 10 most relevant corpora based on a given query from a large corpus. The key challenge is to accurately identify the necessary corpora using a RAG system that incorporates techniques such as embedding and reranking. Task 2 requires generating precise answers to the query based on the corpora retrieved in Task 1. Notably, the corpora may include not only textual data but also numerical tables, and the context sizes of the reference corpora are frequently very large. The critical aspect is to locate the necessary numerical information within the vast corpus and generate accurate responses using LLMs. 

The dataset used in the competition is finance-specific and consists of queries and corpora. Each dataset involves extracting the correct top 10 corpora based on the given query and generating the correct answer. The datasets used are as follows:

\begin{itemize}
\item {\textbf{FinDER}}: 10-K Reports and financial disclosures, to evaluate understanding domain-specific professionals, jargon, and abbreviations.
\item{\textbf{FinQABench}}: 10-K Reports, focusing on detecting hallucinations and ensuring factual correctness in the generated responses.
\item{\textbf{FinanceBench}}: 10-K Reports, to evaluate how well systems handle straightforward, real-world financial questions.
\item{\textbf{TATQA}}: Financial Reports, involving numerical reasoning over hybrid data (tabular and text) to evaluate basic arithmetic, comparisons, and logical reasoning.
\item{\textbf{FinQA}}: Earnings Reports, involving tabular and text data to evaluate multi-step numerical reasoning
\item{\textbf{ConvFinQA}}: Earnings Reports, involving tabular and text data to evaluate handling conversational queries.
\item{\textbf{MultiHiertt}}: Annual Reports, involving hierarchical tables and unstructured text to evaluate complex reasoning tasks involving multiple steps across various document sections.
\end{itemize}

\section{Method and Results}

Our proposed RAG system comprises three main stages. The first two stages focus on the algorithm for Task 1, while the third stage addresses the algorithm for Task 2. Initially, a pre-retrieval process prepares the data for the retrieval stage. In the retrieval stage, we extract a preliminary set of relevant corpora using reranker models, followed by a second reranking to extract more precise corpora. Finally, based on the extracted corpora, the response generation stage produces the final answer.

\subsection{Pre-Retrieval}

The queries in the dataset are typically composed of simple sentences; however, they often include financial-specific abbreviations, ambiguous meanings, or require multi-step reasoning to resolve. To address these challenges, we opted not to use the raw queries directly. Instead, we enhanced them through query expansion techniques \cite{koo2024optimizing, patel2024hypothetical, gao2024modular} aimed at clarifying sentence meanings, interpreting abbreviations, and decomposing complex queries into simpler, more manageable steps. These enhancements were conducted using the {\textit{OpenAI/GPT-4o-mini}}. We employed several query expansion methods, including paraphrasing, keyword extraction, and the creation of hypothetical documents. By combining the results of these methods with the original queries, we aimed to improve the effectiveness of the retrieval process. 

Additionally, to handle the extensive corpus efficiently, we utilized summarization and table extraction. For summarization, we used the {\textit{OpenAI/GPT-4o}} to create condensed summaries of the corpus, which were subsequently used for retrieval. Table extraction was specifically applied to the MultiHiertt dataset, as this dataset contained corpora with extremely large token counts, and much of the critical information was embedded within tables. We conducted an ablation study using various combinations of query expansion techniques and corpus optimizations, with the results presented in Table \ref{pre_retrieval_table}.

Based on these findings, we opted to combine the original query with keyword extraction for all datasets and applied corpus table extraction specifically to the MultiHiertt dataset to optimize retrieval performance.

\begin{table*}[h]
\centering
\setlength{\tabcolsep}{13pt} 
\caption{Ablation study of Pre-Retrieval}
\label{pre_retrieval_table}
\begin{tabular}{cccccccc}
\toprule
\multicolumn{4}{c}{\textbf{Query}} & \multicolumn{3}{c}{\textbf{Corpus}} & \multirow{2}{*}{\textbf{NDCG@10}} \\
\cmidrule(lr){1-4} \cmidrule(lr){5-7}
Original & Paraphrased & \makecell{Keywords\\extraction} & \makecell{Hypothetical\\documents} & Original & Summary& \makecell{Table\\only\textsuperscript{*}} & \\
\midrule
$\checkmark$ & - & - & - & $\checkmark$ & - & - & 0.48949 \\
$\checkmark$ & $\checkmark$ & - & - & $\checkmark$ & - & - & 0.51228 \\
$\checkmark$ & - & $\checkmark$ & - & $\checkmark$ & - & - & 0.54090 \\
$\checkmark$ & - & - & $\checkmark$ & $\checkmark$ & - & - & 0.43707 \\
\midrule
$\checkmark$ & - & - & - & - & $\checkmark$ & - & 0.45589 \\
$\checkmark$ & $\checkmark$ & - & - & - & $\checkmark$ & - & 0.48495 \\
$\textbf{\checkmark}$ & - & $\textbf{\checkmark}$ & - & - & $\textbf{\checkmark}$ & - & 0.48949 \\
$\checkmark$ & - & - & $\checkmark$ & - & $\checkmark$ & - & 0.43707 \\
\midrule
$\checkmark$ & - & - & - & - & - & $\checkmark$ & 0.51228 \\
$\checkmark$ & $\checkmark$ & - & - & - & - & $\checkmark$ & 0.55602 \\
$\textbf{\checkmark}$ & - & $\textbf{\checkmark}$ & - & - & - & $\textbf{\checkmark}$ & \textbf{0.58102} \\
$\checkmark$ & - & - & $\checkmark$ & - & - & $\checkmark$ & 0.45589 \\
\bottomrule
\end{tabular}
\begin{tablenotes}
\item[*]\hspace*{-\tabcolsep}\textsuperscript{\dag} Reranker: \textit{jina-reranker-v2-base-multilingual}
\hfill\textsuperscript{*} Applied only to the MultiHiertt dataset
\end{tablenotes}
\end{table*}

\subsection{Retrieval}

For the retrieval process, we utilized the preprocessed queries and corpora from the pre-retrieval stage. To maximize performance, we directly employed reranker models instead of relying on embedding similarity comparisons. Reranker models offer superior performance as they perform binary classification by processing both the query and the corpus simultaneously.

We initially extracted the top 200 relevant corpora for each query-corpus pair using a relatively lightweight reranker model (\textit{jina-reranker-v2-base-multilingual}) \cite{jina2024reranker}. Subsequently, these top 200 corpora were reranked using more precise reranker models, ultimately selecting the top 10 most relevant corpora. 

To identify the more refined reranker models, we conducted extensive experiments using various top-ranking models from the MTEB leaderboard as well as open-source models. Specifically, we utilized the labels provided by the competition organizers to determine the best-performing reranker model for each dataset. Based on these evaluations, we selected \textit{jina-reranker-v2-base-multilingual}, \textit{gte-multilingual-reranker-base} \cite{alibaba2024gte}, and \textit{bge-reranker-v2-m3} \cite{baai2024bge}, as summarized in Table \ref{retrieval_table}. Through this process, we achieved a final NDCG@10 score of 0.63996 for public leaderboard on Task 1 .

\begin{table*}[htbp]
\centering
\caption{Reranker Models Used for Each Dataset}
\label{retrieval_table}
\setlength{\tabcolsep}{10pt} 
\renewcommand{\arraystretch}{1.2} 
\begin{tabular}{lcccc}
\toprule
\textbf{Dataset} & \multicolumn{3}{c}{\textbf{Reranker}} \\
\cmidrule(lr){2-4}
 & \textbf{jina-reranker-v2-base-multilingual} & \textbf{gte-multilingual-reranker-base} & \textbf{bge-reranker-v2-m3} \\
\midrule
FinDER & - & - & \checkmark \\
FinQABench & \checkmark & - & - \\
FinanceBench & \checkmark & - & - \\
TATQA & - & - & \checkmark \\
FinQA & - & \checkmark & - \\
ConvFinQA & - & - & \checkmark \\
MultiHiertt & \checkmark & - & - \\
\bottomrule
\end{tabular}
\end{table*}

\subsection{Geneartion}
The context size of LLMs has increased exponentially in recent times, with some models capable of handling contexts exceeding 2 million tokens \cite{google2024vertexai}. However, studies \cite{jin2024long, databricks2024rag} have shown that LLM performance degrades as the context size increases. Therefore, selecting an appropriate context size is crucial for optimal performance.

According to recent research \cite{databricks2024rag}, models like \textit{OpenAI/o1-preview} exhibit high performance up to a context size of 64k tokens, beyond which performance declines. In our manual analysis of the competition data, we observed significant performance degradation beyond 32k tokens. Consequently, we utilized up to the top 20 corpora retrieved to limit the context size.

If the input token size (query plus top 1-20 corpora) was under 32k tokens, we processed it directly. If it exceeded 32k tokens, we split the corpora into halves, processed them separately through the LLM, and then fused the resulting answers.

Another critical consideration during generation was the answer format. Financial experts often require specific and concise numerical values rather than lengthy, complex explanations. Therefore, we performed prompt engineering to ensure that the LLM focused on providing the key information and values requested in the query. The overall process of the proposed method is presented in Algorithm \ref{financerag_alg}.

\begin{algorithm}
\caption{Proposed FinanceRAG System}
\label{financerag_alg}
\begin{algorithmic}[1]

\State \textbf{Input:} Query $Q$, Corpus $C$
\State \textbf{Output:} Response $R$

\State $Q' \gets Q + \text{Extracted Keywords from } Q$
\State $C' \gets \text{Extract tables from MultiHiertt, retain other corpus as is}$

\State $C_{\text{top 200}} \gets \text{1st Reranker}(Q', C')$
\State $C_{\text{top 20}} \gets \text{2nd Reranker}(Q', C_{\text{top 200}})$

\If{$\text{Token count of } (Q' + C_{\text{top 20}}) \leq 32k$}
    \State $R \gets \text{LLM}(Q', C_{\text{top 20}})$
\Else
    \State $R_1 \gets \text{LLM}(Q', C_{\text{top 1-10}})$
    \State $R_2 \gets \text{LLM}(Q', C_{\text{top 11-20}})$
    \State $R \gets \text{Fusion}(R_1, R_2)$
\EndIf
    \State \textbf{return} \text{Response} $R$

\end{algorithmic}
\end{algorithm}

\section{Discussion}

While the competition rules prohibited training models using the source data, future work could involve leveraging the source data for model fine-tuning to achieve better reranking performance. Specifically, parameter-efficient fine-tuning (PEFT) techniques could be applied to train only the final layer (e.g., the scoring layer) of the reranker models, thereby enhancing performance on the subtasks.

Moreover, incorporating simple client inputs, such as specifying the search scope (e.g., "APPLE Stock"), could significantly reduce the corpus size, enabling the construction of an accurate RAG system at a lower computational cost. Proper planning and system design can make the solution more efficient and practical for real-world applications.

A notable challenge we encountered was the substantial computational cost and time required. Although financial investment experts often deal with high-value queries where such costs may be justifiable, optimization remains essential. The retrieval and reranking components can be optimized through quantization techniques. However, the LLM input, consisting of the query and corpora, still poses challenges due to its size.

We experimented with using pre-summarized corpora to reduce input size, but this approach led to a significant drop in performance for queries requiring specific numerical responses. Addressing this limitation remains an open area for future research.

\section{Conclusion}

The analysis of financial statements, disclosures, press releases, and related materials is a task of high market value and one that LLMs and RAG systems are well-suited to perform. Developing a finance-specific RAG system entails handling large-scale corpora and requires domain-specific knowledge and numerical data analysis capabilities. By leveraging hybrid embedding similarity functions, long context management techniques, and comprehensive ablation studies on query expansions, we successfully developed a high-performance finance-specific RAG system. We ultimately achieve 2nd place in the FinanceRAG Challenge.

Our work demonstrates that with careful system design and optimization, LLMs can effectively process and analyze complex financial data to generate accurate and valuable insights. We hope that our contributions will further the adoption of AI technologies in the financial industry and inspire future research in this area.

\begin{acks}
We would like to thank the Finance Security Institute (FSI) for the support.
\end{acks}

\bibliographystyle{ACM-Reference-Format}
\bibliography{sample-base}

\end{document}